\documentclass{article}

\PassOptionsToPackage{numbers, compress}{natbib}

\usepackage[preprint]{neurips_2024}

\usepackage[utf8]{inputenc} 
\usepackage[T1]{fontenc}    
\usepackage{hyperref}       
\usepackage{url}            
\usepackage{booktabs}       
\usepackage{amsfonts}       
\usepackage{nicefrac}       
\usepackage{microtype}      
\usepackage{xcolor}         

\usepackage{caption}
\usepackage{graphicx}
\usepackage{booktabs}
\usepackage{todonotes}

\usepackage{amsmath}
\usepackage{enumitem}

\title{Extended Mind Transformers}
\author{%
  Phoebe Klett\thanks{Correspondence to: \texttt{phoebe@normalcomputing.ai}}\\
  Normal Computing \\
  \And
  Thomas Ahle\\
  Normal Computing \\
}

\begin{document}
\maketitle

\begin{abstract}
Pre-trained language models demonstrate general intelligence and common sense, but long inputs quickly become a bottleneck for memorizing information at inference time. We resurface a simple method, Memorizing Transformers (Wu et al., 2022), that gives the model access to a bank of pre-computed memories. We show that it is possible to fix many of the shortcomings of the original method, such as the need for fine-tuning, by critically assessing how positional encodings should be updated for the keys and values retrieved. This intuitive method uses the model's own key/query system to select and attend to the most relevant memories at each generation step, rather than using external embeddings. We demonstrate the importance of external information being retrieved in a majority of decoder layers, contrary to previous work. We open source \footnote[1]{Code and data: \url{https://github.com/normal-computing/extended-mind-transformers/}} a new counterfactual long-range retrieval benchmark, and show that Extended Mind Transformers outperform today's state of the art by 6\% on average. 
\end{abstract}

\section{Introduction and History of Memory Augmented Neural Networks}
Today’s large language models (LLMs) are impressive general learners. During training they memorize large vocabularies, grammars, and wide-ranging knowledge. \cite{brown2020language} In many application settings, however, the model also needs to make use of specialized or topical information. The facts seen during pre-training may have changed, or new information may need to be memorized. Keeping track of the state of the world as the model is queried is crucial for sound reasoning, but efficiently loading a detailed perspective of the world remains a challenge.

This problem can largely be split into the following related but distinct sub-problems: $(1)$ Improve performance by extending the maximum input token sequence length $(2)$ Improve the efficiency of the attention mechanism $(3)$ Improve performance by memorizing or retrieving the relevant information. Longer input sequences allow us to memorize more information at query time, 
improving the efficiency of the attention computation allows us to attend to more information, and retrieval methods allow us to determine which information from the past is relevant to the current query. Many solutions to the problem of leveraging a history of information or long input sequences address more than one of the above sub-problems, and solutions to all three are well-explored \cite{Biggs2015NeuralMN}. We describe their history here to situate our contributions.

\begin{figure}[t]
    \centering
    \includegraphics[width=\textwidth]{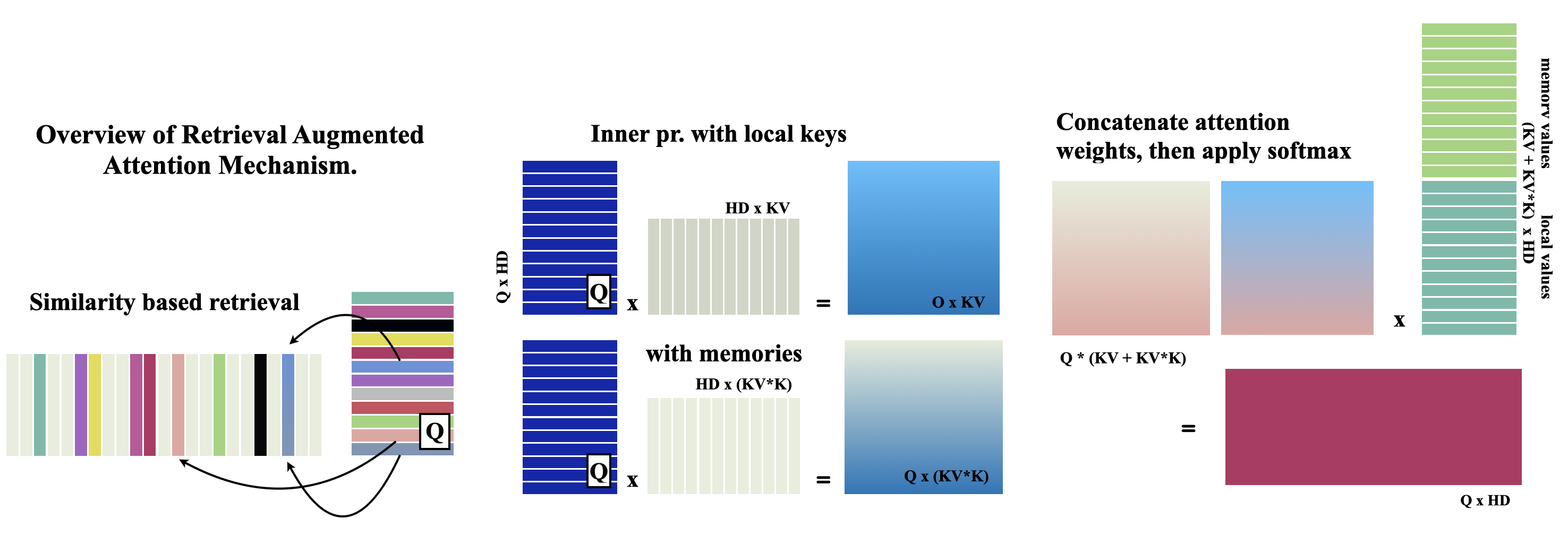}
    \caption{Overview of attention over memories and local context, where Q is length of queries, KV is length of key-values, HD is head dimension, and K is a memory hyper-parameter. Arrows show queries retrieving memories, gradient colors represent inner product scores and softmax.}
    \label{fig:overview}
\end{figure}

\subsection{Extending sequence length}
Auto-regressive language models are trained on sequences of a fixed length. Following the influential introduction of the transformer model in \cite{vaswani2023attention}, Transformer-XL (Dai et al., 2019) \cite{dai2019transformerxl} proposed to generalize beyond that fixed context length by conditioning each time step's hidden states on the previous step's hidden states. In particular, the previous time step's hidden states are concatenated to the current time step's hidden states before computing the key and value states. Importantly, the authors in \cite{dai2019transformerxl} also introduce the first relative position embedding, a crucial part of generalizing beyond training sequence length. Because transformers are position agnostic \cite{yun2020transformers}, vectors encoding position information are added (or otherwise incorporated) to the input embeddings. Absolute position vectors are unique per position, whereas relative position vectors simply encode which tokens are close (far). Compressive transformers \cite{rae2019compressive} built on this work by compressing those hidden states furthest away, those that would fall out of the Transformer-XL \cite{dai2019transformerxl} cache.

Today, strategies to extend the context window have been enabled by improvements in position encodings. Attention with Linear Biases (ALiBi) \cite{press2022train} uses linear biases to damp down the relative importance of long-range tokens rather than using explicit position information. Rotary position embeddings \cite{su2023roformer} generalize the intuition of using the angle between two-dimensional vectors as a distance metric. By interpolating between positions, we can create more unique rotary position embeddings within the expected range \cite{chen2023extending}. Both linear biases and rotary position embeddings enable models to generalize to longer input sequences without fine-tuning, but performance on very long inputs requires additional training.

\subsection{Approximate Attention}
Along a concurrent, parallel path, approximations to the usual causal attention mechanism enabled longer training sequences by reducing computational costs. Sparse attention factorizations of the attention \cite{child2019generating} reduce the quadratic cost of attention \cite{vaswani2023attention} to $O(n \sqrt{n})$. Important drop-in attention approximations (they can be substituted for regular attention without further fine-tuning) were proposed by both Longformer \cite{beltagy2020longformer} and  Memory efficient transformers \cite{gupta2021memoryefficient}. The former introduces a variety of sliding window attention patterns, used now in popular open-sourced LLMs such as Mistral \cite{jiang2023mistral}. The latter proposes top-k attention; attention is only computed for those key-value pairs where the key-query inner product is the largest. This line of research has continued to be fruitful, but it's those hardware-aware improvements \cite{dao2022flashattention} that have become standard. 

\subsection{Retrieval augmentations}
Retrieval methods predate the transformer model. Indeed, many transformer-oriented retrieval methods cite and build on the original ideas proposed in \cite{graves2014neural} Neural turing machines (Graves et al., 2014) and Memory Networks (Weston et al., 2015) \cite{weston2015memory}, which were influential but computationally expensive and difficult to train. Nearest Neighbor Language Models (kNN-LM) \cite{khandelwal2020generalization} interpolate the model's output distribution with a distribution over labels of similar inputs seen during training. KNN Information Fetching Modules\cite{fan2020augmenting} operate by retrieving relevant (potentially multi-modal) external sources and concatenating with the encoded input before feeding into the decoder modules in a classic encoder-decoder transformer. Retrieval-Augmented generation (RAG) (Lewis et al., 2021) \cite{lewis2021retrievalaugmented} builds directly on both the previous ideas and proposes to retrieve external documents based on similarity with the current input sequence, and condition on those documents as additional context during generation. The encoding used to perform retrieval need not be the same encoding used by the seq2seq model, since the tokens are appended to the input sequence as text, previous to any generation. 

All of the aforementioned methods represent query and long-range information using standard embeddings. Much of the recent work on retrieval has explored the optimal way to represent long-range information for retrieval. The Memory Transformer (Burtsev et al., 2021) \cite{burtsev2021memory}, similar to Transformer-XL \cite{dai2019transformerxl}, demonstrates that long-range information can be compressed and memorized in dedicated \texttt{[MEM]} tokens, both using shared and disjoint weight matrices from the query tokens. Finally, Memorizing Transformers (Wu et al., 2022) \cite{wu2022memorizing} proposes to cache previous key-value representations, and perform top-k attention \cite{gupta2021memoryefficient} over those key-value pairs in a single decoder layer. We believe this method has been underestimated, and it forms the basis for our work in this paper.

\begin{figure}[t]
    \centering
    \begin{minipage}{0.48\textwidth}
        \centering
        \captionsetup{font=footnotesize, margin={.5cm, .5cm}} %
        \includegraphics[width=\textwidth]{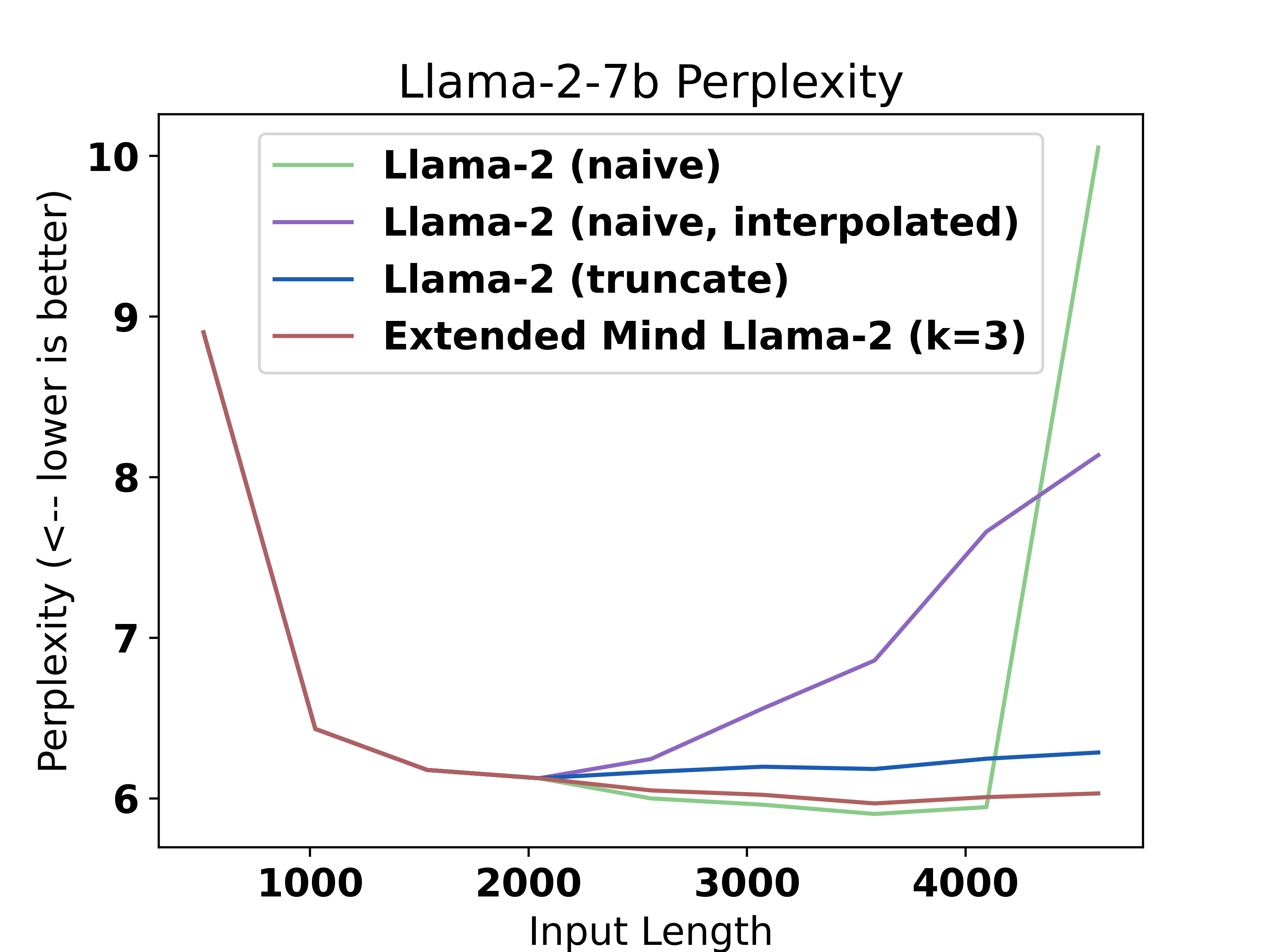} 
        \caption{Average perplexity on sequences of increasing input lengths. Results shown for baselines and Extended Mind Llama-2-7b.}
        \label{fig:perplexity-llama}
    \end{minipage} \hfill
    \begin{minipage}{0.48\textwidth}
        \centering
        \captionsetup{font=footnotesize, margin={.5cm, .5cm}}
        \includegraphics[width=\textwidth]{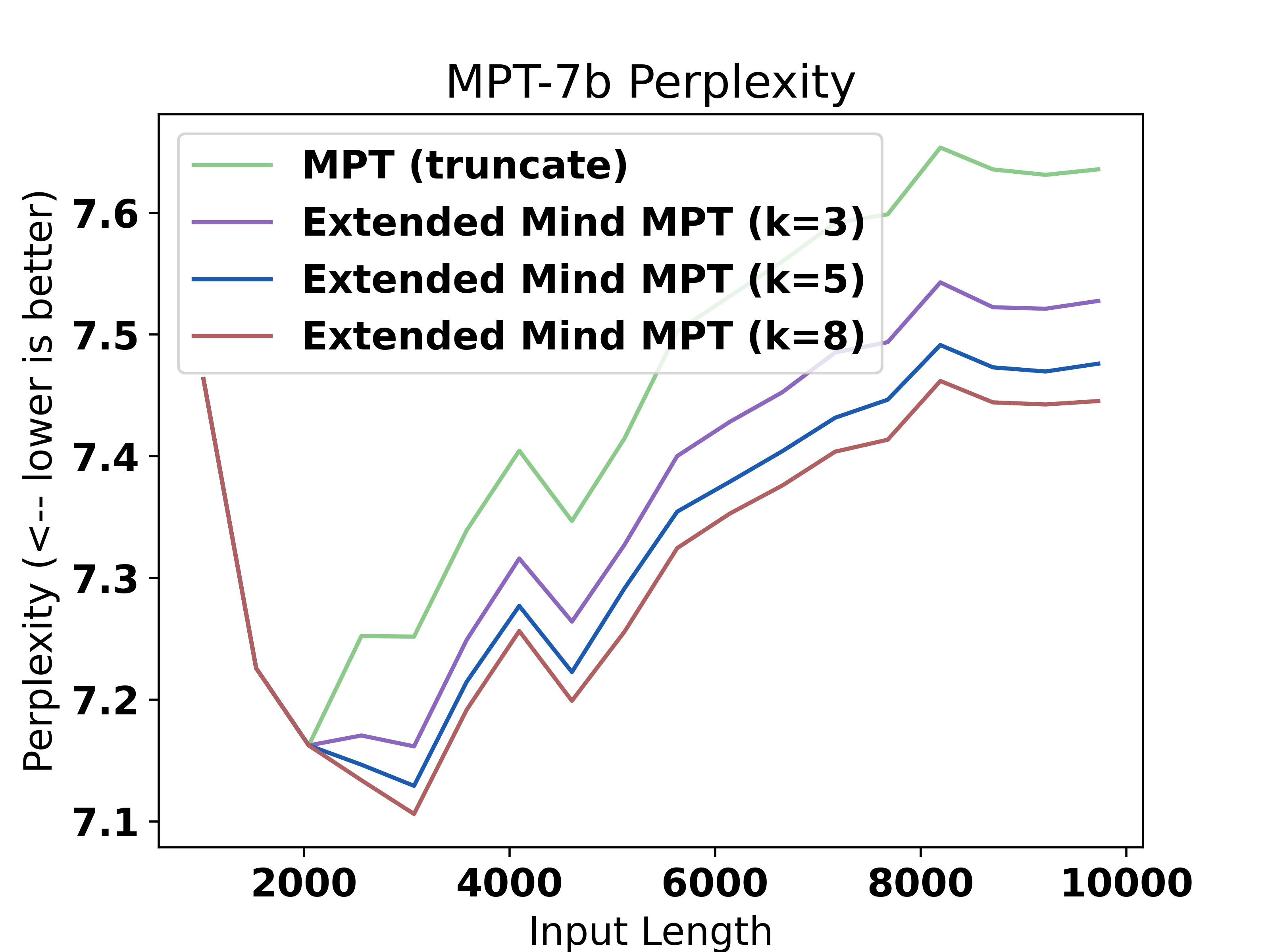} 
        \caption{Average perplexity on sequences of increasing input lengths. Increasing k corresponds to retrieving more key-values pairs.}
        \label{fig:perplexity-mpt}
    \end{minipage}
\end{figure}

\subsection{Extended Mind Transformers}
After so much experimentation, what have we learned? We've traded the recurrent mechanism of caching and conditioning on previous hidden states, as implemented by Transformer-XL \cite{dai2019transformerxl}, for the simplicity and interpretability of similarity-based retrieval methods. We see this learning as analogous to the Transformer advantage over LSTM models. Recent advances in position encodings allow models to generalize to longer inputs more natively, and some approximations to classic attention have made their way into practice. But the two most popular methods, fine-tuning models to memorize long input sequences and RAG \cite{lewis2021retrievalaugmented} both leave much to be desired.

\begin{figure}[t]
    \centering
    \begin{minipage}{0.48\textwidth}
        \centering
        \captionsetup{font=footnotesize, margin={.5cm, .5cm}} %
        \includegraphics[width=\textwidth]{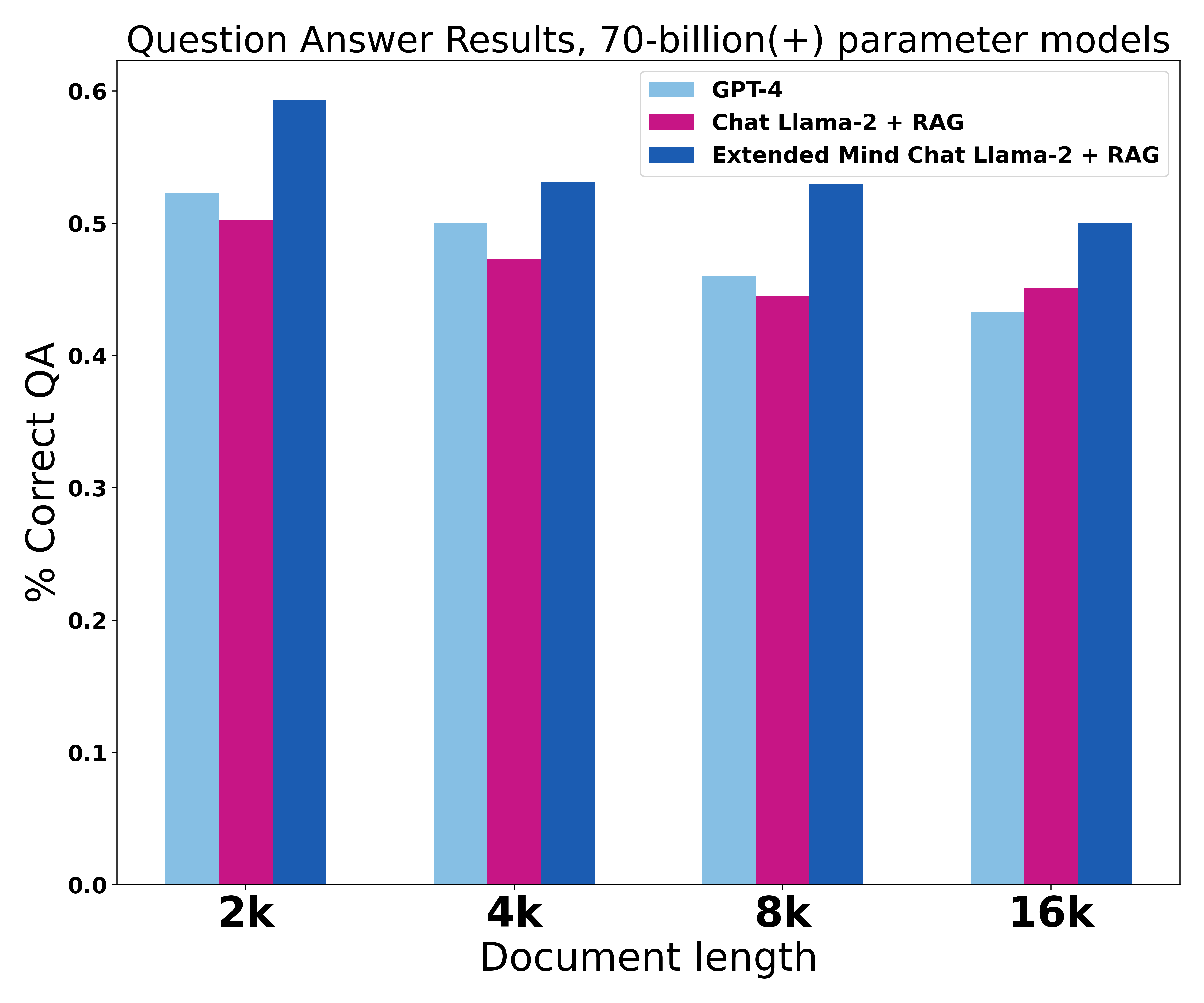} 
        \caption{Fact retrieval accuracy over various document lengths for Extended Mind Llama-2-70b, RAG and state of the art baselines.}
        \label{fig:rag}
    \end{minipage} \hfill
    \begin{minipage}{0.48\textwidth}
        \centering
        \captionsetup{font=footnotesize, margin={.5cm, .5cm}}
        \includegraphics[width=\textwidth]{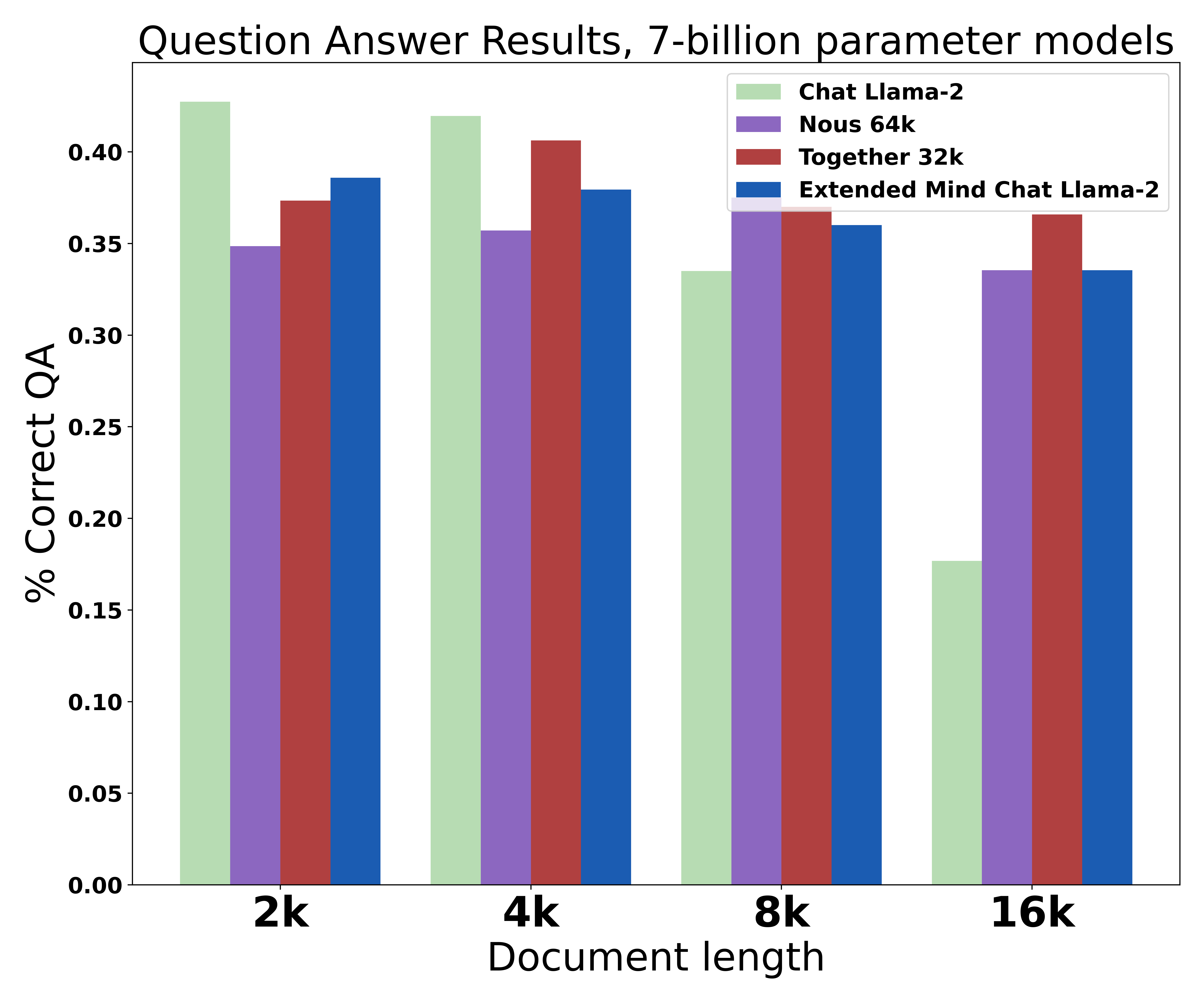} 
        \caption{Retrieval accuracy over various document lengths for Extended Mind Llama-2-7b, and long context baselines.\\}
         \label{fig:ft}
    \end{minipage}
\end{figure}

Putting the entire history into context suffers the quadratic cost of attention unnecessarily. More often than not, only a subset of the new information is needed at any given generation step. Further, our experiments \ref{fig:ft} show that attention over long inputs continues to struggle on long-range retrieval tasks even after fine-tuning. A more optimal solution would combine attention over long(er) inputs with a smart retrieval mechanism. This is achieved today by combining fine-tuned models with RAG. But while RAG \cite{lewis2021retrievalaugmented} has become widespread and easy to implement \cite{llamaindex}, it faces many challenges. All retrieval mechanisms external to the model are fundamentally limited by relying on an external embedding model, rather than the transformer's own key/query system. This decision is made once per query and used for all generation steps. Assuming that the entire long inputs do not fit into context (otherwise RAG would be unnecessary), we cannot pose queries that synthesize over all inputs. Challenges that face both fine-tuning and RAG include recent evidence that language models struggle to ignore irrelevant and false information \cite{chen2023benchmarking}. 

Retrieval methods internal to the model avoid these issues by retrieving only the most relevant information at each step, where relevance is measured using the model's representation of the data. They have yet to be adopted in practice due to the staleness problem, or the fear that representations will become out of distribution as the model weights evolve, and because the additional complexity and overhead of fine-tuning is difficult to justify when RAG is often effective. We hypothesized that just as the relative position embeddings enable native generalization to longer inputs, they should enable models to use past key-values retrieved within decoder layers, as proposed by \cite{wu2022memorizing}. If true, this would eliminate both the need to fine-tune and thus the staleness problem as well. 

\subsection{Our Contributions}
In this work we make the following contributions:
\begin{itemize}[noitemsep, topsep=1pt, parsep=1pt, partopsep=1pt, leftmargin=20pt]
    \item We propose \textbf{Extended Mind Transformers}, a variety of decoder-only transformers closely related to Memorizing Transformers \cite{wu2022memorizing} that retrieve and attend to an external cache of key-value pairs (or memories) without finetuning. 
    \item We demonstrate that both models trained with ALiBi and rotary position embeddings can leverage retrieved information natively, and present the hyper-parameter decisions that enable this. Namely, we show $(1)$ how to add position information to retrieved key-value pairs and $(2)$ that the retrieved information must be accessible to a majority of decoder layers.
    \item We open-source a new counterfactual long-range retrieval dataset, and demonstrate that Extended Mind Transformers outperform today's SoTA models.
    \item We demonstrate that Extended Mind Transformers are time-efficient. The upfront cost of generating the cached key-values pairs is quickly amortized so our method is a low cost performance boost for tasks that query the same documents many times.
    \item We demonstrate the benefit of causal citations and new active learning generation paradigms enabled by Extended Mind Transformers. 
    \end{itemize}

\section{Methods} 

We introduce a simple kNN-based retrieval mechanism into the decoder-only transformer architecture, just as in \cite{wu2022memorizing}. This enables the model to attend to a select number of external memories during each generation step. The model decides for each token, within a particular decoder, which memories are important. We describe the process of generating the external memories, the addition of top-k attention \cite{gupta2021memoryefficient} into the attention module, and the important choices of which layers to augment with this retrieval mechanism, as well as how to assign position information to external memories. 

\subsection{Generating external memories}
Given a set of long inputs, generating the external memories amounts to passing all inputs through the model and caching its internal representations of the data. This mechanism is already built into the model's architecture to speed up generation; we don’t need to recompute these values for previously seen tokens during the generation process. We compute these representations using a stride of length $s$ so that each token is conditioned on at least $s$ tokens. Figure~\ref{fig:timing} shows how we can balance precision and efficiency by tuning the stride length. Smaller strides generate higher-quality representations, while larger strides require fewer computations.  We generate the cache once, and documents may be processed in parallel. The upfront cost for each document is $n$ passes through the model with $L$ number of input tokens, where $n$ is the smallest integer such that $(sn) + L \geq$ document length.

The cost of attention at inference time (over all documents) is $O(d(n_{q}(n_{q} + k)) + n_{q} * RT(n_{D}, k))$ where $n_{q}$ is the length of the query, $n_{D}$ is the length of the long inputs, $k$ is the number of retrieved memories (per query token), $d$ is the dimension of the representation, and $RT$ is a choice of $topk$ retrieval function (could be an approximate or exact method). Classic self-attention over all documents costs $O(d(n_{d}+n_{q})^{2})$ for the same inference query. When $D$ is large, we realize time-savings which is visualized in an inference times experiment in Figure ~\ref{fig:timing}.

\subsection{Extended Mind Attention}
The retrieval-augmentation is quite straightforward. In addition to self-attention, we introduce top-k attention \cite{gupta2021memoryefficient} over the cached key-value pairs. I.e., we allow each query token to attend to a certain number of external key-value pairs where the cosine similarity between the query and key is largest. We use cosine similarity rather than inner product to retrieve key-value pairs to mitigate distribution shift, following \cite{wu2022memorizing}. Explicitly, the retrieval-augmented attention mechanism can be described by the following equations:
\begin{align}
    \operatorname{scores} =
    \frac{K_{R}}{\mid \mid K_{R} \mid \mid} \times \frac{Q}{\mid \mid Q \mid \mid} \hspace{.5cm},\hspace{.5cm}
  \break
    \operatorname{softmax}\left(\frac{Q(K_{M}\oplus K_{L})^{T}}{\sqrt{d}}\right) \times \left(V_{M} \oplus V_{L}\right)
\end{align}
where $\oplus$ refers to tensor concatenation, $(K_{L}, V_{L})$ are key-value pairs from local context, $(K_{R}, V_{R})$ are all key-value pairs from external memories, and $(K_{M}, V_{M})$ are those $k$ key-value pairs with largest score. As usual, $Q$ are queries, and $d$ is head dimension. Figure~\ref{fig:overview} provides a visual aid for how self-attention is combined with top-k attention over external memories. 

We mask the attention weights such that each query token can only attend to its own retrieved keys, and not those retrieved by previous or following query tokens. This could enable even cheaper computations, but we do not implement it in our code. We find this masking does not diminish performance as compared to causal masking methods. To further speed up the retrieval computations we can use a vector database designed exactly for this purpose. We support using FAISS \cite{douze2024faiss} in our implementation. 

\subsection{Augmenting a subset of layers}
Although previous work \cite{wu2022memorizing} suggests using retrieval-augmentation methods on a subset or single decoder layer, we found it crucial to use our top-k augmentation on a majority of decoder layers. As proposed by Clark and Chalmers in the influential work ``The Extended Mind'' \cite{emh}, the external information must be constantly and immediately accessible in order to successfully contribute to the model's memory. We name our class of models after this pioneering work, and provide evidence for this claim in the experiments we describe below. In the experiments that follow, we show results in Appendix \ref{sec:appendix-ppl} for Extended Mind Transformers that use memory-augmented attention on only the last half or third of the decoders. These results, especially for retrieval tasks, are quite poor. Although not explicitly discussed in \cite{wu2022memorizing}, we believe it's likely the cost of fine-tuning was a main barrier to using more than one memory-augmented layer in previous work. 

\subsection{Position Information}
As hypothesized, we find that both models trained with ALiBi and rotary position embeddings can use retrieved key-value pairs natively. The position information assigned to those retrieved key-value pairs is dependent on the position encoding used. 

For models trained with ALiBi, such as MPT \cite{MosaicML2023Introducing}, we found that placing the cache at a position just after the inputs was effective. In this case, the model interprets the retrieved memories as being a constant distance (one position) away from the tokens it considers local context. For models trained using rotary position embeddings, such as Llama2 \cite{touvron2023llama}, we found that using no position information at all on retrieved key-value pairs was most effective. Whereas all the local tokens are rotated to encode their distance from other tokens, tokens retrieved from the memory are not rotated at all. 

\subsection{Pruning Memories}
Retrieving too many memories can cause generations to decrease in quality. We implement two simple pruning techniques to keep generations high quality while retrieving many memories to maximize recall. This can be seen as regularization.

In the first method, we mask retrieved tokens that don't meet some low similarity threshold. We found $0.25$ to be a good choice. We implement this for both model architectures we open source, but find it particularly important for models trained with ALiBi. This can dampen retrieval capabilities, however. Key-query similarity scores even for useful tokens can be surprisingly low and even negative in models trained with rotary position embeddings.

In contrast, we found our second pruning technique to be crucial for models trained with rotary positions, and less so for those trained with ALiBi. This method removes memories that correspond to special tokens; tokens that encode the beginning or end of a sequence or an unknown token. We remove these tokens after memories are generated, rather than removing the tokens from the history. More detail on both regularization techniques can be found in Appendix ~\ref{sec:appendix-reg}, and each experiment's corresponding Appendix section describes which techniques were used.

\section{Experiments}
We perform three types of experiments: we run perplexity experiments to measure how models trained with ALiBi and rotary position embeddings are able to leverage the memory tokens during next-token prediction, we run counterfactual retrieval experiments to measure the model's ability to recall and report facts in the external memory, and we run inference-time experiments to measure the time saving of using Extended Mind Transformers over regular attention. All code for reproducing experiment results can be found in the attached code and will be open-sourced.

\begin{figure}[t]
    \centering
    \includegraphics[width=0.60\textwidth]{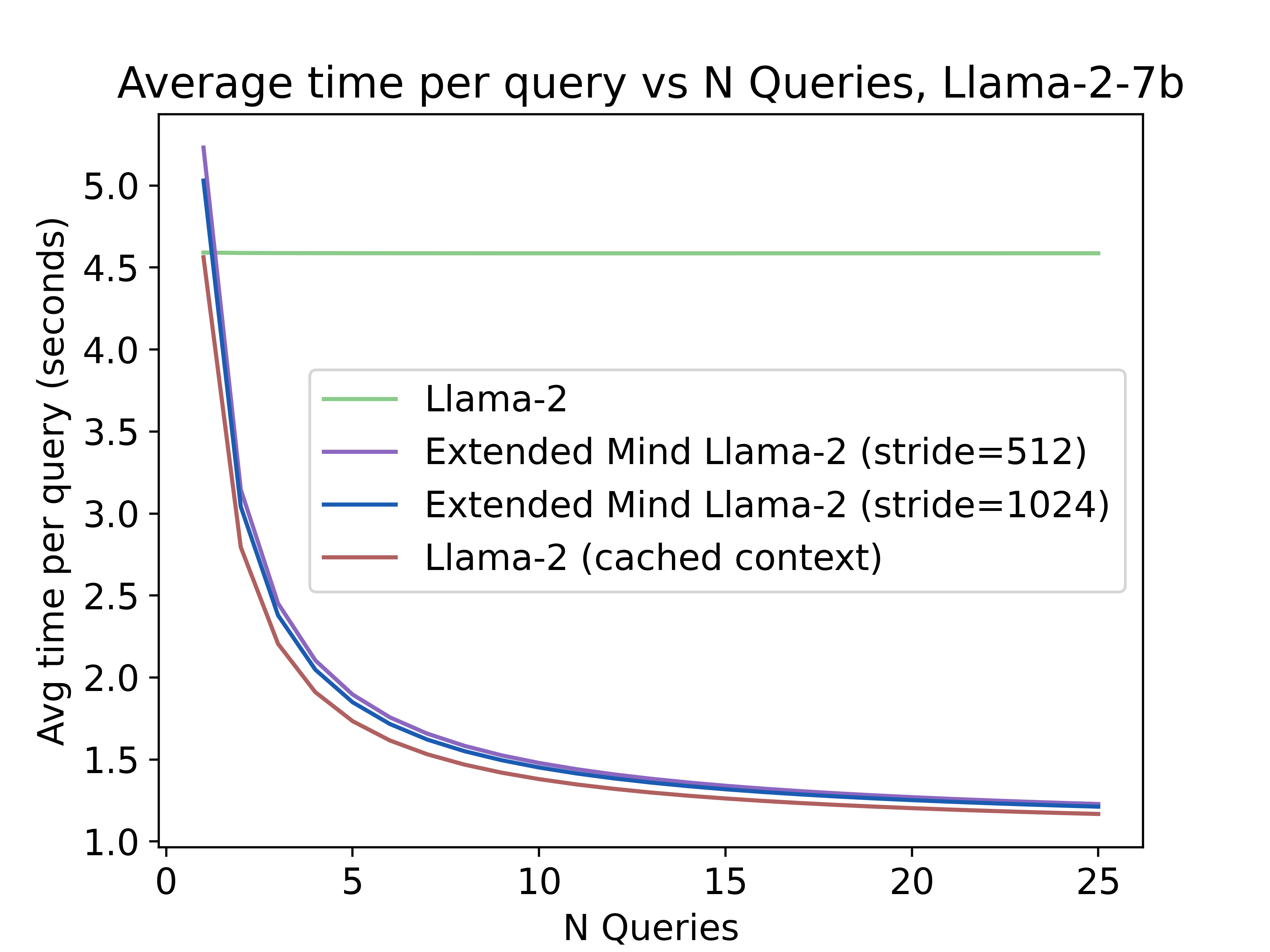}
    \caption{Average inference time for many queries over documents of length 4000 tokens.}
    \label{fig:timing}
\end{figure}
\subsection{Perplexity Experiments}
We use perplexity as a metric for overall model performance. Perplexity is a measure of uncertainty over each generated token, related to the cross-entropy loss function. We batch documents into sequences of increasing input lengths, and compute perplexity using a stride of $512$ tokens. We use the  Wikitext-103 dataset \cite{merity2016pointer}. Detail on batching can be found in Appendix ~\ref{sec:appendix-ppl}.

We provide multiple baselines to compare against Extended Mind Transformers. The naive baseline sees all inputs in context, regardless of their length. The truncate baseline only sees the last $N$ tokens in context, where $N$ is the sequence length the model saw during training. Extended Mind Transformers see the last $N$ tokens in context (the same as the truncate baseline), but also attend to the previous tokens as external memories. We only compute perplexity on the last $N$ tokens for all methods. We run the experiment for two 7-billion parameter models, MPT-7b \cite{MosaicML2023Introducing} and Llama-2-7b \cite{touvron2023llama}. In the case of Llama-2-7b, we compute naive baseline methods both with and without interpolated position embeddings. Results for these experiments are shown in Figures~\ref{fig:perplexity-llama}, ~\ref{fig:perplexity-mpt} and above.

\begin{table}
    \caption{Perplexity Results for a sample of input lengths.}
    \centering
    \label{table:perplexity}
        \begin{tabular}{lccccc}
            \toprule
            Model & \hspace{-1.5cm} Input length: & 2560 & 4608 & 6656 & 8704 \\
            \midrule
            Extended Mind Llama2-7B (k=3) &&  6.091 & 6.031 & 6.062 & 6.143 \\
            Llama2-7B (truncate) && 6.207 & 6.286 & 6.383 & 6.477 \\
            Llama2-7B (naive) &&  5.100 &  10.052 & 49.657 & n/a \\
            Llama2-7B (naive, interpolated) &&  6.245 &  8.137 & 9.807 & n/a \\
            Extended Mind MPT-7B (k=8) && 7.134 & 7.199 & 7.376 & 7.444\\
            MPT-7B (truncate) && 7.252 & 7.347 & 7.560 & 7.636 \\
            MPT-7B (naive) && 7.233 & 23.273 & n/a &  n/a \\
            \bottomrule
        \end{tabular}
\end{table}

The naive baseline demonstrates the phenomenon methods to extend the input token sequence length seek to ameliorate: after sequences exceed lengths greater than 2-3k tokens, the performance quickly drops off. In this case, perplexity quickly blows up. Using interpolated position embeddings staves off the drop temporarily, but not past input sequence lengths greater than around 6k.

Both Extended Mind Transformers, MPT-7b, and Llama-2-7b are able to leverage retrieved key-value pairs without any fine-tuning, providing strong evidence that a variety of relative positional encodings enable this capability. Extended Mind Transformers outperform both baselines. Even more, perplexity continues to decrease as we increase the number of retrieved key-value pairs per query token for models trained with ALiBi. We also provide results in Appendix ~\ref{sec:appendix-ppl} for Extended Mind Llama-2-7b where only a subset of decoder layers retrieve key-value pairs from the cache. These results are significantly worse than retrieving on all decoder layers. 

\subsection{Counterfactual Retrieval Experiments}
We propose that Extended Mind Transformers address both the issue of extending the maximum input token sequence length (the memory cache is unlimited) and integrating retrieval into generation. Thus, we measure the quality of the retrieval augmented attention in Extended Mind Transformers with that of pre-trained models, fine-tuned models, and composite RAG methods. Extended Mind Transformers are just as compatible with RAG as other models, and we report these results as well. We found MPT models to be much less competitive on this benchmark, and omit performance metrics.

We proxy the quality of attention with a challenging counterfactual retrieval task. Our dataset is a modified wikiQA benchmark \cite{abacusLongContext}, which we provide along with our code. The dataset is composed of Wikipedia articles (of 2-16 thousand tokens) and corresponding questions. We modify the dataset by changing the labeled answers to realistic but wrong answers, to control for facts memorized during pre-training. For example, we replace every instance of ``Lee Hazlewood'' with ``Terry Allen'' in the Wikipedia entry for the song ``These Boots Were Made For Walking'', and then ask the model to produce the songwriter's name, with the correct answer now being ``Terry Allen''. Details for dataset creation can be found in Appendix ~\ref{sec:appendix-retrieval}. 

We run two sizes of Extended Mind Transformers on our new benchmark, based on the 7-billion and 70-billion parameter Chat Llama-2 \cite{touvron2023llama} models. No Extended Mind Transformers have been further fine-tuned. For comparators to the 7-billion parameter model we run Chat Llama-2-7b with interpolated position embeddings (with and without RAG), as well as two base Llama-2-7b models that have been fine-tuned on longer input sequences. ``Yarn-Llama-2-7b-64k'' \cite{peng2023yarn} (also instruct fine-tuned) and ``LLaMA-2-7B-32K''\cite{together} were developed by Nous Research and Together Computer, respectively. As a comparator to the 70-billion parameter model, we test the popular closed-source model GPT-4 \cite{openai2024gpt4}. Details on the RAG implementation can be found in Appendix \ref{sec:appendix-retrieval}.

All comparator models see the entire prompt and document in context. RAG methods select the best $\approx 2500$ tokens from the document to include in context. Extended Mind Transformers see only the prompt and question in context, and attend to the document only through the retrieval-augmented attention. When combined with RAG, they see the best $\approx 2500$ tokens in context, and attend to the document in memory. Details on prompting can be found in Appendix ~\ref{sec:appendix-retrieval}, as well as results for Extended Mind Chat Llama-2-7b where only some decoders retrieve key-value pairs from the cache.

\begin{table}
    \caption{Retrieval results for all models.}
    \label{tab:retrieval}
    \centering
        \begin{tabular}{lccccc}
            \toprule
            Model & \hspace{1.5cm} & 2k & 4k & 8k & 16k \\
            \midrule
            Extended Mind Chat Llama2-7b &&  38.59\% & 37.95\% & 36.00 \% & 33.54\% \\
            Chat Llama2 && 42.74\% & 41.96\% & 33.50\% & 17.68\% \\
            Nous && 34.85\% & 35.71\% & 37.50\% & 33.54\% \\
            Together AI& & 37.34\% & 40.63\% & 37.00\% & 36.59\% \\
            \midrule
            Extended Mind Chat Llama2-70b + RAG && 59.34\% & 53.13\% & 53.00\% & 50.00\% \\
            Chat Llama2-70b + RAG &&  50.21\% & 47.32\% & 44.50\% & 45.12\% \\
            GPT-4 && 52.28\% & 50.00\% & 46.00\% & 43.29\% \\ 
            \bottomrule
        \end{tabular}
\end{table}
Figure \ref{fig:ft} shows the overall retrieval accuracies of Extended Chat Llama-2-7b as compared to other methods for extending input length. Extended Mind Chat Llama outperforms both fine-tuned models on short inputs, and the baseline model on long inputs, competitive with the fine-tuned models. Figure \ref{fig:rag} shows the overall retrieval accuracies of Extended Chat Llama-2-70b as compared to other retrieval methods. While neither RAG nor our top-k retrieval alone are enough to beat GPT-4 across the board, Extended Mind Chat Llama combined with RAG improves upon GPT-4 by a large margin, $6\%$ when averaged across input lengths. We also include more detailed heatmap visuals in Appendix \ref{sec:appendix-retrieval} that introduce the added dimension of how many times a retrieved fact appeared in the document. 

\subsection{Inference Times}
We demonstrate the time-efficiency of Extended Mind Transformers by plotting average inference time for queries over documents of 4k tokens. Extended Mind Transformers incur an upfront cost to generate the memories, and then need only compute self-attention on the query and retrieved memories. We compare this to both a naive and more sophisticated benchmark. The naive method (this is the only option for closed-source models) puts the document into context alongside the query, for all queries. Hence the constant inference time, as shown in Figure~\ref{fig:timing}. As a more sophisticated comparison, we cache the key-value pairs generated for the document after the first query and avoid re-computing them each successive query. 

We show performance for Extended Mind Transformers that use a stride length equal to both 512 (default) and 1024 (recommended for efficiency) when generating memories. Further detail on hyper-parameters can be found in the  Appendix ~\ref{sec:appendix-timing}. While Extended-Mind models do incur the additional upfront cost, this is quickly amortized over additional queries, making our performance improvement a low cost improvement. 

\section{Interpretability and Impact to Reasoning}

In this final section, we discuss a new kind of citation and active learning generation technique enabled by Extended Mind Transformers. 
\begin{figure}
    \centering
    \includegraphics[width=0.85\textwidth]{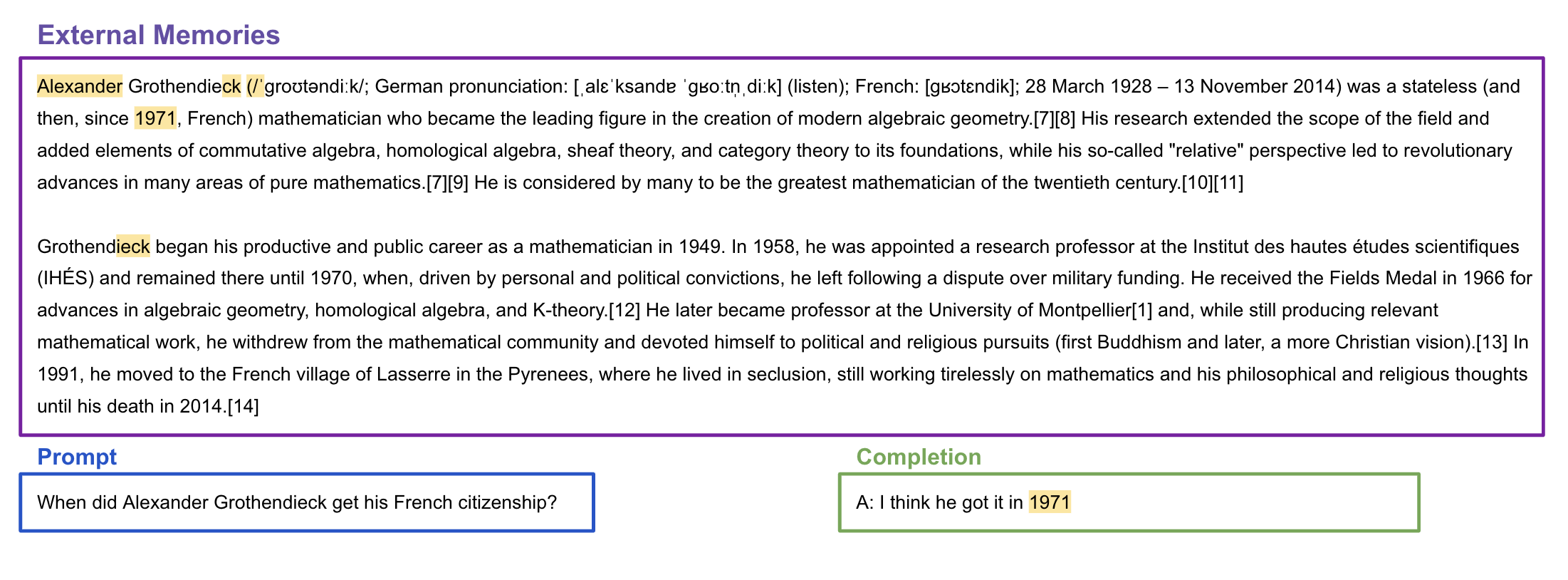}
    \caption{Highlighted tokens are those most retrieved memories when generating the token ``1971''.}
    \label{fig:explain}
\end{figure}

\subsection{Citations}
Beyond performance improvements, Extended Mind Transformers also enable a new kind of causal citations. Local attention weights are difficult to interpret \cite{jain2019attention} due to their scale. Simply reporting similar in-context information or which information from external documents was included in context (for instance, as retrieved by RAG) has no guaranteed causal relation to what information was used during generation. Instead, we can now easily report which memories were retrieved and attended to at each step of the generation. This gives us a much better idea of which information the model used to generate each next token. We enable this in our open-source models, returning the indices of retrieved memories along with model outputs. Figure~\ref{fig:explain} demonstrates an example of this. Since we retrieve key-value pairs within each layer, we make some standard aggregation choices that are detailed further in Appendix ~\ref{sec:appendix-citations}. 

\subsection{Uncertainty and Active Learning}
Extended Mind Transformers also enable new insight into when a model is uncertain about its generation. When a model is uncertain about its generation, or equivalently it didn't memorize the required information during training, its generations evolve as we allow the model to retrieve more information from the cache. Conversely, if the model was sure of its answer with a small amount of retrieved information, adding new memories will not change its output. We provide examples in Appendix ~\ref{sec:appendix-activelearning}. 

\begin{center}
\includegraphics[width=0.85\textwidth]{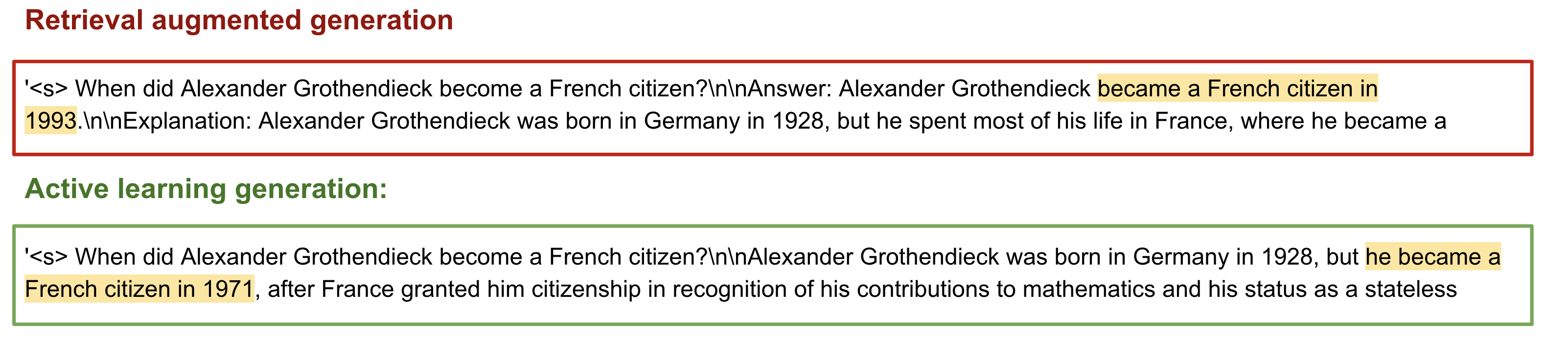}
\label{fig:rag2}
\captionof{figure}{Example of using active learning generation with Extended Mind Chat Llama-2.}
\end{center}

More than this, Extended Mind Transformers suggest a new way to reduce hallucinations. Using simple uncertainty metrics, i.e. token level entropy, we can detect when the model is uncertain about its generated token, and allow the model to regenerate the step using more information from the cache. We leave full-scale experiments for this technique to future papers, but provide a demonstrative example in Figure~\ref{fig:rag2}, and further details in \ref{sec:appendix-activelearning}. We suspect this active learning aware generation will also improve given more robust uncertainty metrics. We are particularly interested in using second-order uncertainty from language models that have been fine-tuned to be Bayesian. 

\section{Conclusion}
We provide an overview of the history of Memory augmented neural networks, and situate our new method, Extended Mind Transformers in this rich history. We propose a new variety of transformer models that retrieve and attend to external documents without finetuning, based on the work in \cite{wu2022memorizing}. Extended Mind Transformers retrieve cached key-value pairs within each decoder layer, for each generated token. We demonstrate that relative positional embeddings allow today's pretrained LLMs to use retrieved key-value pairs without fine-tuning, eliminating the previous barriers to using retrieval mechanisms internal to language models. We also demonstrate the importance of the external information being constantly and immediately accessible by comparing with models which only use retrieval on a subset of decoder layers. 

We open-source a new counterfactual long-range retrieval benchmark, and show that Extended Mind Transformers are competitive with models that have been fine-tuned on long inputs, despite not being fine-tuned, and provide new SoTA results when combined with RAG. We measure and report the time savings of using Extended Mind Transformers when making multiple queries over long documents. Finally, we discuss the new kind of citations enabled by Extended Mind Transformers, as well as the potential for new active learning inspired generation techniques. 

\bibliographystyle{abbrvnat}
\bibliography{references}

\section{Appendix}

All code and instructions for reproducing experiment results can be found in the accompanying code.

\subsection{Experiment Details: Perplexity}
\label{sec:appendix-ppl}

\textbf{Resources used:} (2x) A100-80GB GPUs, $3-24$ hours, depending on method and parameters. 

\textbf{Additional Experiment Settings:}
\begin{enumerate}
    \item All models use $N = 2048$. I.e. we cut off (or cache) inputs longer than 2048 tokens for truncate (memorizing) runs.
    \item Llama baseline with interpolation uses dynamic position interpolation with an $\alpha=8$, which is standard.
    \item We do not use similarity masking \ref{sec:appendix-reg} in any perplexity experiments. 
    \item We do remove memories associated to unknown and special tokens \ref{sec:appendix-reg} for Llama models. 
    \item The below graphic illustrates how different methods handle input sequences, batched documents of subsequences of increasing lengths. For simplicity, the graphic shows a stride equal to the input length. 
\end{enumerate}

\begin{center}
\includegraphics[width=0.85\textwidth]{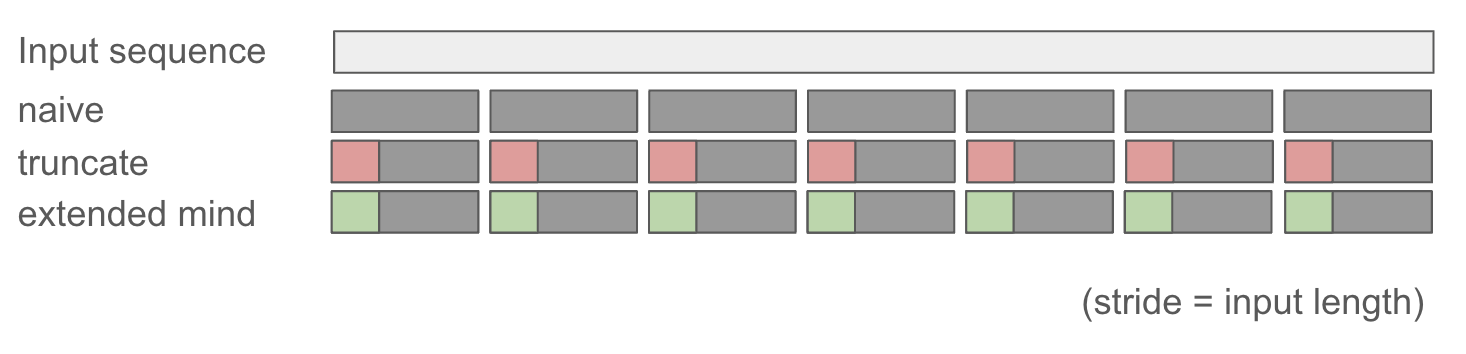}
\end{center}

\textbf{Additional Results}
Perplexity results for models with retrieval on only a subset (the last third and half) of decoder layers:
\begin{center}
    \label{tab:ppl-lessmem}
    \begin{tabular}{lccccc}
        \toprule
        Model & \hspace{-1.5cm} Input length: & 2560 & 4608 & 6144 \\
        \midrule
        Extended Mind Llama2-7B (retrieval on last third) &&  6.181 & 6.212 & 6.271 \\
        Extended Mind Llama2-7B (retrieval on last third) && 6.162 & 6.167 & 6.212 \\
        \bottomrule
    \end{tabular}
\end{center}

\subsection{Experiment Details: Counterfactual Retrieval}
\label{sec:appendix-retrieval}

\textbf{Resources used:} (2-4x) A100-80GB GPUs, $3-24$ hours, depending on method and parameters. 

\textbf{Heatmap visuals}
The below figures illustrate the percent of correctly retrieved facts by bucketed number of fact appearances. Each cell represents samples where the number of fact appearances falls within the range specified on the x-axis, and is colored to represent the overall success of the retrievals. The results are further stratified by sequence length, on the y-axis. Since we did not create the dataset from scratch, this isn't consistent from document to document. Intuitively, if the fact appears many times, it should be easier to retrieve.
\vspace{.3cm}
\begin{figure}
    \centering
    \includegraphics[width=\textwidth]{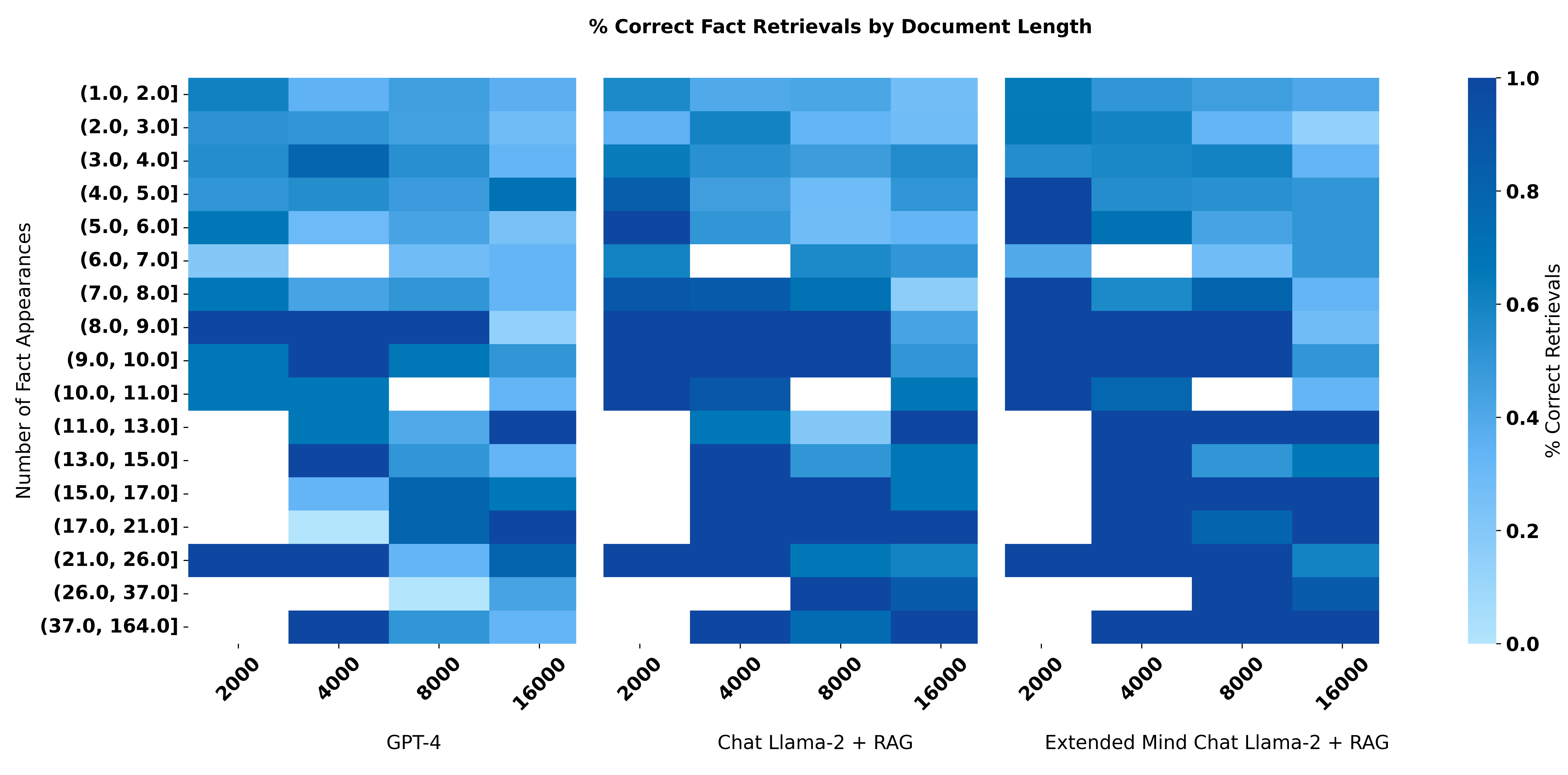}
    \caption{Retrieval results Base Model, Fine-tuned Models, and Extended Mind Transformer. Cells represent a bucket (x-axis), and sequence length (y-axis). Colors to represent the overall success of the retrievals.}
    \includegraphics[width=\textwidth]{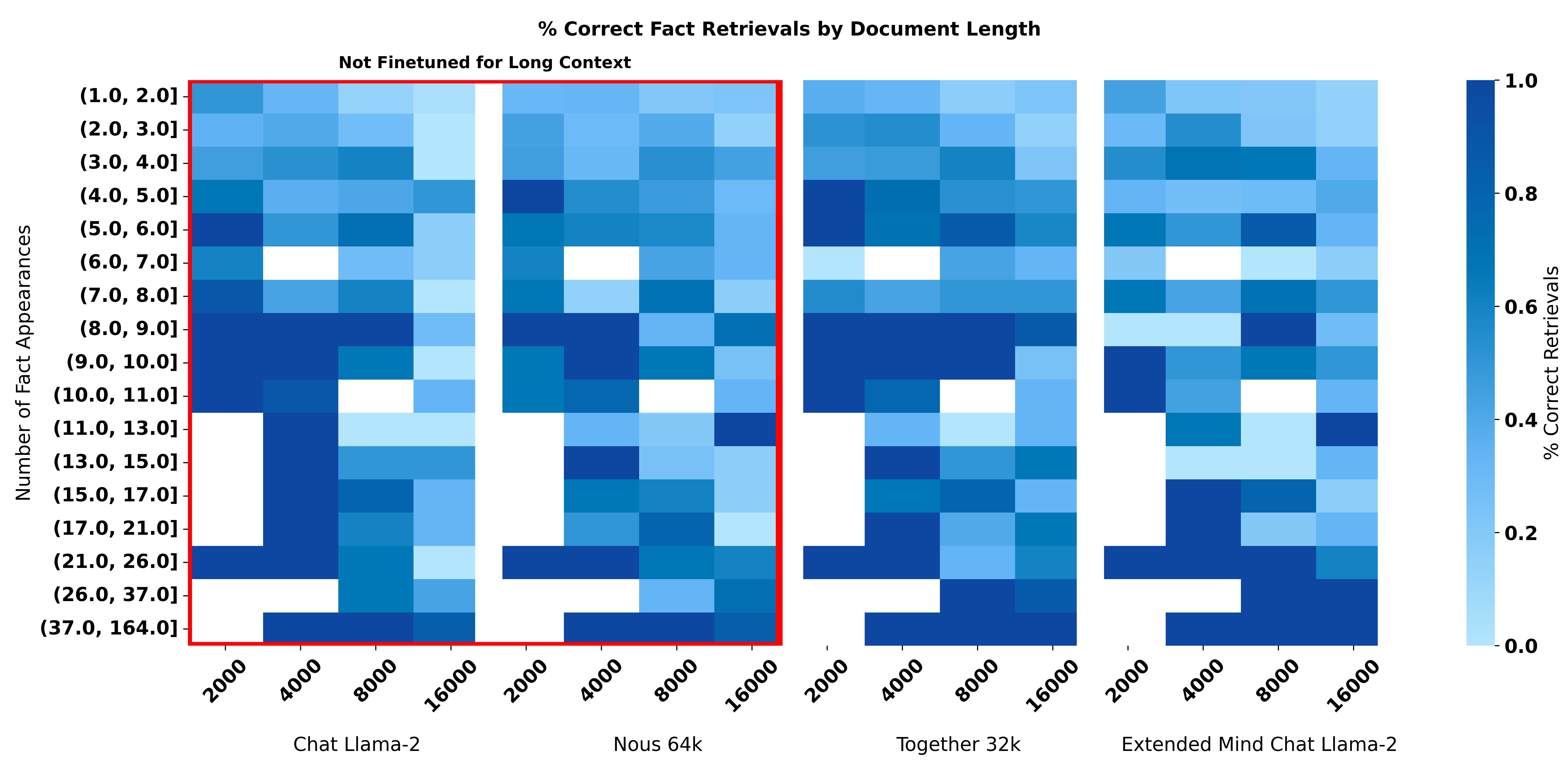}
    \caption{Retrieval results Base Model, Fine-tuned Models, and Extended Mind Transformer. Cells represent a bucket (x-axis), and sequence length (y-axis). Colors to represent the overall success of the retrievals.}
    \label{fig:heatmap}
\end{figure}

\textbf{Dataset Creation:}

Substitutions were made in a few ways. 
\begin{enumerate}
    \item Substitution of an entity with a semantically similar entity. E.g. England vs. France, or Hank Aaron vs Joe Adcock.
    \item Substitution of an entity with a syntactically similar entity. E.g. Beatrix vs Beatrice.
    \item Number / Date substitution. E.g. 1990 vs 1999, or 30 ml vs. 100 ml.
\end{enumerate}

\textbf{Additional Experiment Settings:}
\begin{enumerate}
    \item Chat models are prompted using the appropriate assistant syntax, and models that haven't been instruction fine-tuned are given a one-shot example. Extended Mind Transformers are prompted using one-shot example, no matter the base model. This is all included in the code. 
    \item Extended Mind Llama-2-7b uses a $k$ value of 12; each query token is allowed to retrieve 12 keys from memory. Extended Mind Llama-2-70b uses a $k$ value of 4.
    \item We use a stride length of 512 when generating the memory cache, and allow 3072 input tokens. While the models saw sequences of 2048 tokens during training, perplexity experiments confirm that input lengths up to 4096 are comprehensible.
    \item We implement a simple RAG mechanism. We chunk the documents into token sequences of length 500 with no overlap, and embed them using OpenAI's Ada model. We then compare similarity to the embedded query, and retrieve the top-5 chunks (about 2500 tokens). 
    \item There are two natural ways to combine RAG with Extended Mind Transformers. $(1)$ Put whatever cannot fit in context into memory, or $(2)$ Put everything into memory. While we observe benefit from both, the results reported reflect the second strategy.
    \item We do not use similarity masking \ref{sec:appendix-reg} in any retrieval experiments. 
    \item Baseline models use dynamic position interpolation with $\alpha=8$.
    \item We do remove memories associated to unknown and special tokens \ref{sec:appendix-reg} for Extended Mind Llama models. 
\end{enumerate}

\textbf{Additional Results}
Perplexity results for models with retrieval on only a subset (the last third and half) of decoder layers. Results are total number correct, rather than percentage, since recall is so low. Datapoint counts for each split are listed alongside.
\begin{center}
     \begin{tabular}{lccccc}
            \toprule
            Model & \hspace{1.75cm} & 2k & 4k & 8k & 16k \\
            \midrule
            Extended Mind Llama2-7B (retrieval on last third) &&  5 & 9 & 5 & 6\\
            Extended Mind Llama2-7B (retrieval on last half) &&  5 & 7 & 5 & 4 \\
            \bottomrule
        \end{tabular}
\end{center}
Datapoint counts: 241 (2k), 224 (4k), 200 (8k), 164 (16k)

\subsection{Experiment Details: Inference Times}
\label{sec:appendix-timing}

\textbf{Resources used:} (2x) A100-80GB GPUs, $1-3$ hours, depending on method and parameters. 

\textbf{Additional Experiment Settings:}
\begin{enumerate}
    \item We use base Llama-2-7b for these experiments.
    \item We allow inputs up to 4096 tokens when generating the memories. While the model saw sequence of length 2048 during training, the model can comprehend sequences up to twice that length, as shown in our perplexity experiments.
\end{enumerate}

\subsection{Further Discussion: Pruning Memories}
\label{sec:appendix-reg}

\textbf{When to use similarity masking:}
We found similarity masking useful for both models trained with ALiBi and rotary position embeddings when doing long form text generation. However, we did not find it necessary on short retrieval tasks for the Llama models. 

\textbf{When to remove special tokens:}
We found it crucial to remove special tokens when retrieving memories from raw web documents, especially those with HTML remnants. This prevents unknown tokens from being retrieved and confusing the model. While removing these tokens does not appear to have much downside, we did not find it as necessary for models trained with ALiBi.

\subsection{Further Discussion: Citations}
\label{sec:appendix-citations}

Since we retrieve key-value pairs within each layer, we simply report those tokens which correspond to the key value pairs which were most retrieved throughout the entire model. 

It would be interesting to do a further analysis on which tokens are retrieved by layers earlier / later in the decoder stack, we leave this to future work. 

\subsection{Further Discussion: Active Learning}
\label{sec:appendix-activelearning}

Figure \ref{fig:rag2} uses Chat Llama-2-7b. The original generation retrieves using a constant $k=1$, and the Active learning generation uses a baseline constant $k=1$, and allows tokens with an entropy above $0.60$ to regenerate using an additional $2$ memories, $k=3$ in total.

Evolution of uncertain and certain generations. This example uses MPT-7b with default parameters, and $k$ specified as below. All settings can be found in the code, and are enumerated below as well.

\textbf{Experiment Settings}
\begin{enumerate}
    \item Model used is MPT-7b \cite{MosaicML2023Introducing}.
    \item We do use similarity masking, with a threshold of $0.25$.
    \item We use $k$ as specified below.
\end{enumerate}

\textbf{Uncertain Example}\\
Prompt: ``When did Alexander Grothendieck get his French citizenship?''\\
Baseline Generation:  ``When did Alexander Grothendieck get his French citizenship?
I am trying to find out when Alexander Grothendieck got his French citizenship. I know that he was born in Germany and that he got his French citizenship in the late 1950s. I am trying to find out when he''\\
Generation for k=5:  ``A: I think he got it in the early 1960s''\\
Generation for k=6:  ``A: I think he got it in the early 1970s.''\\
Generation for k=7:  ``A: He was born in France, and he was naturalized in 1971.
<|endoftext|>''\\
Generation for k=8:  ``A: I think he got it in 1971.
<|endoftext|>Q''\\
Generation for k=30:  ``A: He was born in Germany, and became a French citizen in 1971.''\\

\textbf{Certain Example}\\
Prompt: ``What was did Alexander Grothendieck's profession?''\\
Baseline Generation:  ``What was did Alexander Grothendieck's profession?
Alexander Grothendieck was a French mathematician'' \\
Generation for k=2:  ``Alexander Grothendieck was a mathematician.$\backslash$n$\backslash$n What'' \\
Generation for k=8:  ``A: He was a mathematician.
<|endoftext|>Q: What'' \\

\end{document}